\documentclass[12pt]{article} 
\usepackage{times}  
\usepackage{helvet} 
\usepackage{courier} 
\usepackage{url}   
\usepackage{natbib}   
\usepackage{fullpage}   
\usepackage{graphicx}
\usepackage{subcaption}
\usepackage{float}

\usepackage{booktabs}    
\usepackage{microtype}   
\usepackage{nicefrac}    
\usepackage{array}
\newcolumntype{P}[1]{>{\centering\arraybackslash}p{#1}}
\usepackage[usenames, dvipsnames]{color}
\usepackage{todonotes}

\usepackage{amsfonts}    
\usepackage{amsmath,amssymb} 
\usepackage{mathtools}

\usepackage{algorithm}
\usepackage[noend]{algpseudocode}

\usepackage{graphicx}
\graphicspath{ {images/} }
\usepackage{authblk}

\title{An Evaluation of the Human-Interpretability of Explanation}

\author[1]{Isaac Lage*}
\author[1]{Emily Chen*}
\author[1]{Jeffrey He*}
\author[1]{Menaka Narayanan*}
\author[2]{Been Kim} 
\author[1]{Sam Gershman} 
\author[1]{Finale Doshi-Velez} 
\affil[1]{Harvard University}
\affil[2]{Google Brain}

\date{\vspace{-5ex}}

\begin{document}

\maketitle

\begin{abstract}
Recent years have seen a boom in interest in machine learning systems
that can provide a human-understandable rationale for their
predictions or decisions. However, exactly what kinds of explanation
are truly human-interpretable remains poorly understood. This work
advances our understanding of what makes explanations interpretable
under three specific tasks that users may perform with machine
learning systems: simulation of the response, verification of a
suggested response, and determining whether the correctness of a
suggested response changes under a change to the inputs. Through
carefully controlled human-subject experiments, we identify
regularizers that can be used to optimize for the interpretability of
machine learning systems. Our results show that the type of complexity 
matters: cognitive chunks (newly defined concepts) affect performance 
more than variable repetitions, and these trends are consistent across 
tasks and domains. This suggests that there may exist some common design 
principles for explanation systems. 
\end{abstract}

\section{Introduction}
\label{sec:intro}

Interpretable machine learning systems provide not only decisions or
predictions but also explanation for their outputs. Explanations can
help increase trust and safety by identifying when the recommendation
is reasonable and when it is not. While interpretability has a long
history in AI \citep{michie1988machine}, the relatively recent
widespread adoption of machine learning systems in real, complex
environments has lead to an increased attention to interpretable
machine learning systems, with applications including understanding
notifications on mobile devices
\citep{mehrotra2017interpretable,wang2016bayesian}, calculating stroke
risk \citep{letham2015interpretable}, and designing materials
\citep{raccuglia2016machine}. Techniques for ascertaining the
provenance of a prediction are also popular within the machine
learning community as ways for us to simply understand our
increasingly complex models
\citep{lei2016rationalizing,selvaraju2016grad,adler2016auditing}.

The increased interest in interpretability has resulted in many forms
of explanation being proposed, ranging from classical approaches
such as decision trees \citep{breiman1984classification} to input
gradients or other forms of (possibly smoothed) sensitivity analysis
\citep{selvaraju2016grad,ribeiro2016should,lei2016rationalizing},
generalized additive models \citep{caruana2015intelligible},
procedures \citep{singh2016programs}, falling rule lists
\citep{wang2015falling}, exemplars
\citep{kim2014bayesian,frey2007clustering} and decision sets
\citep{lakkaraju2016interpretable}---to name a few. In all of these
cases, there is a face-validity to the proposed form of explanation:
if the explanation was not human-interpretable, clearly it would not
have passed peer review.

That said, these works provide little guidance about when different
kinds of explanation might be appropriate, and within a class of
explanations---such as decision-trees or decision-sets---what factors most
influence the ability of humans to reason about the explanation. 
For example, it is hard to imagine that a human would find a 
5000-node decision tree as interpretable as a 5-node decision tree for any 
reasonable notion of interpretable, but it is not clear whether it is more 
urgent to regularize for fewer nodes of shorter average path lengths. 
In \citet{doshi2017roadmap}, we point to a growing need for the
interpretable machine learning community to engage with the human
factors and cognitive science of interpretability: we can spend
enormous efforts optimizing all kinds of models and regularizers, but
that effort is only worthwhile if those models and regularizers
actually solve the original human-centered task of providing
explanation.

Determining what kinds of regularizers we should be using, and when,
require carefully controlled human-subject experiments. In this work,
we make modest but concrete strides towards providing an empirical
grounding for what kinds of explanations humans can utilize. Focusing
on decision sets, we determine how three different kinds of
explanation complexity---clause and explanation lengths, number and
presentation of cognitive chunks (newly defined concepts), and variable
repetitions---affect the ability of humans to use those explanations
across three different tasks, two different domains, and three
different performance metrics. We find that the type of complexity 
matters: cognitive chunks affect performance more than variable repetitions, 
and these trends are consistent across tasks and domains. This suggests 
that there may exist some common design principles for explanation systems. 

\section{Related Work}

\textbf{Interpretable Machine Learning} Interpretable machine learning
methods aim to optimize models for both succinct explanation and
predictive performance. Common types of explanation include
regressions with simple, human-simulatable functions
\citep{caruana2015intelligible, kim2015ibcm,
 ruping2006learning,buciluǎ2006model, ustun2016supersparse,
 doshi2014graph, kim2015mind, krakovna2016increasing,
 hughes2016supervised, jung2017simple}, various kinds of logic-based
methods
\citep{wang2015falling,lakkaraju2016interpretable,singh2016programs,liu2016sparse,safavian1991survey,
 bayesian2017wang}, techniques for extracting local explanations from
black-box models
\citep{ribeiro2016should,lei2016rationalizing,adler2016auditing,selvaraju2016grad,
 smilkov2017smoothgrad,shrikumar2016not,kindermans2017patternnet,
 ross2017right}, and visualization \citep{wattenberg2016attacking}.
There exist a range of technical approaches to derive each form of
explanation, whether it be learning sparse models
\citep{mehmood2012review,chandrashekar2014survey}, monotone functions
\citep{canini2016fast}, or efficient logic-based models
\citep{rivest1987learning}. Related to our work, there also exists a
history of identifying human-relevant concepts from data, including
disentangled representations \citep{chen2016infogan} and predicate
invention in inductive logic programming \citep{muggleton2015meta}.
While the algorithms are sophisticated, the measures of
interpretability are often not---it is common for researchers to
simply appeal to the face-validity of the results that they find
(i.e., ``this result makes sense to the human reader'')
\citep{caruana2015intelligible,lei2016rationalizing,ribeiro2016should}.

\textbf{Human Factors in Explanation} In parallel, the literature on
explanation in psychology also offers several general insights into
the design of interpretable AI systems. For example, humans prefer
explanations that are both simple and highly probable
\citep{lombrozo07}. Human explanations typically appeal to causal
structure \citep{lombrozo2006structure} and counterfactuals
\citep{keil2006explanation}. \citet{miller1956magical} famously
argued that humans can hold about seven items simultaneously in
working memory, suggesting that human-interpretable explanations
should obey some kind of capacity limit (importantly, these items can
correspond to complex \emph{cognitive chunks}---for example,
`CIAFBINSA' is easier to remember when it is recoded as `CIA', `FBI',
`NSA.'). Orthogonally, \citet{kahneman2011thinking} notes that humans
have different modes of thinking, and larger explanations might push
humans into a more careful, rational thinking mode. Machine learning
researchers can convert these concepts into notions such as sparsity
or simulatability, but answering questions such
as ``how sparse?'' or ``how long?'' requires empirical evaluation in
the context of machine learning explanations.

\textbf{A/B Testing for Interpretable ML} Existing studies evaluting
the human-interpretability of explanation often fall into the A-B test
framework, in which a proposed model is being compared to some
competitor, generally on an intrinsic task. For example,
\citet{kim2014bayesian} showed that human subjects' performance on a
classification task was better when using examples as representation
than when using non-example-based
representation. \citet{lakkaraju2016interpretable} performed a user
study in which they found subjects are faster and more accurate at
describing local decision boundaries based on decision sets rather
than rule lists. \citet{subramanian1992comparison} found that users
prefer decision trees to tables in games, whereas
\citet{huysmans2011empirical} found users prefer, and are more
accurate, with decision tables rather than other classifiers in a
credit scoring domain. \citet{hayete2004gotrees} found a preference
for non-oblique splits in decision trees (see
\citet{freitas2014comprehensible} for a more detailed survey). These
works provide quantitative evaluations of the human-interpretability
of explanation, but rarely identify which properties are most
essential for which contexts---which is critical for generalization.

\textbf{Domain Specific Human Factors for Interpretable ML} Specific
application areas have also evaluated the desired properties of an
explanation within the context of the application. For example,
\citet{tintarev2015explaining} provides a survey in the context of
recommendation systems, noting differences between the kind of
explanations that manipulate trust \citep{cosley2003seeing} and the
kind that increase the odds of a good decision
\citep{bilgic2005explaining}. In many cases, these studies are
looking at whether the explanation has an effect on performance on 
a downstream task, sometimes also
considering a few different kinds of explanation (actions of similar
customers, etc.). \citet{horsky2012interface} describe how presenting
the right clinical data alongside a decision support recommendation
can help with adoption and trust. \citet{bussone2015role} found that
overly detailed explanations from clinical decision support systems
enhance trust but also create over-reliance; short or absent
explanations prevent over-reliance but decrease trust. These studies
span a variety of extrinsic tasks, and again given the specificity of
each explanation type, identifying generalizable properties is
challenging.

\textbf{General Human Factors for Interpretable ML} Closer to the
objectives of this work, \citet{kulesza2013too} performed a
qualitative study in which they varied the soundness (nothing but the
truth) and the completeness (the whole truth) of an explanation in a
recommendation system setting. They found completeness was important
for participants to build accurate mental models of the system.
\citet{allahyari2011user,elomaa2017defense} also find that larger
models can sometimes be more interpretable. \citet{schmid2016does}
find that human-recognizable intermediate predicates in inductive
knowledge programs can sometimes improve simulation time.
\citet{sangdeh2017manipulating} manipulate the size and transparency
of an explanation and find that longer explanations and black-box
models are harder to simulate accurately (even given many instances)
on a real-world application predicting housing prices. Our work fits
into this category of empirical study of explanation evaluation; we
perform controlled studies on a pair of synthetic applications to
assess the effect of a large set of explanation parameters.

\section{Methods}
Our central research question is to determine which properties of
decision sets are most relevant for human users to be able to use the
explanations for a set of synthetic tasks described below. In order
to carefully control various properties of the explanation and the
context, we generate explanations by hand that mimic those learned by
machine learning systems. We emphasize that while our explanations
are not machine-generated, our findings provide suggestions to the
designers of interpretable machine learning systems about which
parameters are most urgent to optimize when producing explanations
since they affect usability most heavily.

The question of how humans utilize explanation is broad. For this
study, we focused on explanations in the form of \emph{decision sets}
(also known as rule sets). Decision sets are a particular form of
procedure consisting of a collection of cases, each mapping some
function of the inputs to a collection of outputs. An example of a 
decision set is given below
\begin{figure}[H] 
 \centering
 \includegraphics[width=3in]{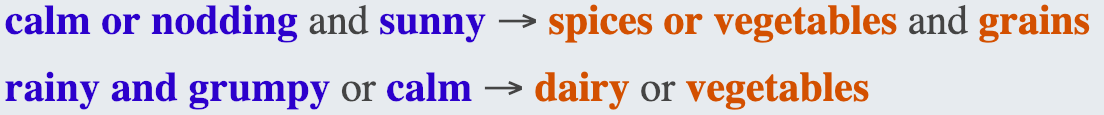}
 \caption{Example of a decision set explanation.}
 \label{fig:rule_set}
\end{figure}
\noindent
where each line contains a clause in disjunctive normal form (an
or-of-ands) of the inputs (blue words), which, if true, maps to 
the output (orange words--also in disjunctive normal form).

Decision sets make a reasonable starting point for a study on
explanation because there exist many techniques to optimize them given
data \citep{frank1998generating, cohen1995fast, clark1991rule,
 lakkaraju2016interpretable}; they are easy for humans to parse since
they can scan for the rule that applies and choose the accompanying
output \citep{lakkaraju2016interpretable}; and there are many
parameters to tune that may influence how easy it is to parse a
specific decision set, including the number of lines, number of times
variables are repeated, and whether terms represent intermediate
concepts. Finally, decision sets can either be trained as the machine
learning model for a given application, or they can be trained to
locally approximate a more complex model like a neural network using
procedures like the one described in \citet{ribeiro2016should}.

Of course, within decision sets there are many possible variations,
tasks, and metrics. Table~\ref{tab:factors} summarizes the
core aspects of our experiments. We considered three main kinds of
explanation variation---variations in explanation size (number of
lines, length of lines), variations in introducing new cognitive
chunks (newly defined concepts), and variations in whether terms repeat. The effect of these
variations were tested across three core cognitive tasks in two
domains, for a total of six settings. We also considered three
metrics---accuracy, response time, and subjective satisfaction---as
measures of task performance that we may care about. In the
following, we first describe each of these core experimental aspects
and then we detail the remaining aspects of the experimental design.

\begin{table}[]
\begin{tabular}{l|l|l|l}
\textbf{Setting: Domain} & \textbf{Setting: Choice of Task} & \textbf{Explanation Variation} & \textbf{Metrics}    \\ \hline
Recipe          & Verification           & V1: Explanation Size      & Response Time      \\
Clinical         & Simulation            & V2: Cognitive Chunks      & Accuracy        \\
             & Counterfactual          & V3: Repeated Terms       & Subjective Satisfaction
\end{tabular}
\caption{We conduct experiments in 2 parallel domains--one low-risk (recipe) and one
  high risk (clinical). In each domain, we conduct 3 experiments testing different 
  types of explanation variation. For each of those, we vary 1-2 factors (described in 
  detail in Section~\ref{sec:exp-variation}). For each setting of these factors, 
  we ask people to perform 3 core cognitive tasks and for each of these, we record 3 
  metrics measuring task performance.}
\label{tab:factors}
\end{table}

\subsection{Setting: Domain}

The question of what kinds of explanation a human can use implies the
presence of a setting, task, and metric that the explanation will
facilitate. Examples include improving safety, where a user might use
the explanation to determine when the machine learning system will
make a mistake; and increasing trust, where a user might be convinced
to use a machine learning system if it justifies its actions in
plausible ways. While evaluating how well explanations facilitate
real world tasks is the ultimate goal, it is challenging because we
must control for the knowledge and assumptions that people bring with
them when making these decisions. For example, even if the task is
the same---improving safety---people may bring different assumptions 
based on aspects of the domain (e.g., perceived risk of the decision).

To rigorously control for outside knowledge that could influence our
results, we created two domains---recommending recipes and medicines to aliens
---to which humans could not bring any
prior knowledge. Further, each question involved a supposedly
different alien to further encourage the users to not generalize
within the experiment. All non-literals (e.g., what ingredients were
spices) were defined in a dictionary so that all participants would
have the same concepts in both experiments. 
Although designed to feel very different, this synthetic set-up also allowed us to
maintain exact parallels between inputs, outputs, categories, and the
forms of explanations.
\begin{itemize}
\item \textbf{Recipe} Study participants were told that the
 machine learning system had studied a group of aliens and determined
 each of their individual food preferences in various settings (e.g.,
 weekend). This scenario represents a setting in which
 customers may wish to inspect product recommendations. This domain
 was designed to feel like a low-risk decision. Here, the inputs are
 settings, the outputs are groups of food, and the recommendations are 
 specific foods.

\item \textbf{Clinical} Study participants were told that the
 machine learning system had studied a group of aliens and determined
 personalized treatment strategies for various symptoms (e.g., sore
 throat). This scenario closely matches a clinical decision support
 setting in which a doctor might wish to inspect the decision support
 system. This domain was designed to feel like a high-risk decision. Here,
 the inputs are symptoms, the outputs are classes of drugs, and the recommendations 
 are specific drugs. We chose drug names that start with the first
 letter of the drug class (e.g., antibiotics were Aerove, Adenon and
 Athoxin) so as to replicate the level of ease and familiarity of
 food names.
\end{itemize}
We hypothesized that the trends would be consistent across both
domains.

\subsection{Setting: Choice of Task}
In any domain, there are many possible tasks that an explanation could
facilitate. For example, one could be interested in error in recipe
suggestions or a mechanism for alien disease. To stay general, as
well as continue to rigorously control for outside knowledge and
assumptions, we follow the suggestion of \citet{doshi2017roadmap} and
consider a core set of three cognitive tasks designed to test how well
humans understand the explanation:
\begin{itemize}
\item \textbf{Simulation} Predicting the system's recommendation given
 an explanation and a set of input observations. Participants were
 given observations about the alien and the alien's preferences and
 were asked to make a recommendation that would satisfy the alien. See
 Figure~\ref{fig:interface_sim}.
\item \textbf{Verification} Verifying whether the system's
 recommendation is consistent given an explanation and a set of input
 observations. Participants were given a recommendation as well as
 the observations and preferences and asked whether it would satisfy
 the alien. See Figure~\ref{fig:interface}.
\item \textbf{Counterfactual} Determining whether the system's
 recommendation changes given an explanation, a set of input
 observations, and a perturbation that changes one dimension of the
 input observations. Participants were given a change to one of the
 observations in addition to the observations, preferences and
 recommendation and asked whether the alien's satisfaction with the
 recommendation would change. See Figure~\ref{fig:interface_ctr}.
\end{itemize}
As before, we hypothesized that while some tasks may be easier or
harder, the trends of what makes a good explanation would be
consistent across tasks. 

\subsection{Explanation Variation}
\label{sec:exp-variation}

While decision sets are interpretable, there are a large number of
ways in which they can be varied that potentially affect how easy they
are for humans to use. Following initial pilot studies (see
Appendix~\ref{sec:pilot}), we chose to focus on three main sources of variation:
\begin{itemize}
\item \textbf{V1: Explanation Size.} We varied the size of the
 explanation across two dimensions: the \emph{total number of lines}
 in the decision set, and the \emph{number of terms within the output
  clause}. The first corresponds to increasing the vertical size of
 the explanation---the number of cases---while the second corresponds
 to increasing the horizontal size of the explanation---the
 complexity of each case. We focused on output clauses because they were harder to parse:
 input clauses could be quickly scanned for setting-related terms,
 but output clauses had to be read through and processed completely
 to determine the correct answer. We hypothesized that increasing
 the size of the explanation across either dimension would increase
 response time. For example, the explanation in Figure~\ref{fig:interface} has 4 lines
 (in addition to the first 3 lines that define what we call explicit cognitive chunks), and
 3 terms in each output clause.

\item \textbf{V2: Cognitive Chunks.} We varied
 the number of cognitive chunks, and whether they were implicitly or
 explicitly defined. We define a cognitive chunk as a clause in 
 disjunctive normal form of the inputs that may recur throughout the decision set. 
 Explicit cognitive chunks are mapped to a name that is then used to reference the 
 chunk throughout the decision set, while implicit cognitive chunks recur throughout the 
 decision set without ever being explicitly named. 
 On one hand, creating new cognitive chunks can
 make an explanation more succinct, while on the other hand, the
 human must now process an additional idea. We hypothesized that
 explicitly introducing cognitive chunks instead of having long
 clauses that implicitly contained them would reduce response time.
 For example, the explanation in Figure~\ref{fig:interface} has 3 explicit cognitive
 chunks, and the explanation in Figure~\ref{fig:interface_ctr} has 3 implicit cognitive
 chunks.
  
\item \textbf{V3: Repeated Terms.} We varied the
 number of times that input conditions were repeated in the decision
 set. If input conditions in the decision list have little overlap,
 then it may be faster to find the appropriate one because there are
 fewer relevant cases to consider than if the input conditions appear
 in each line. Repeated terms was also a factor used by
 \cite{lakkaraju2016interpretable} to measure interpretability. We
 hypothesized that if an input condition appeared in several lines of
 the explanation, this would increase the time it took to search for
 the correct rule in the explanation. For example, each of the
 observations in Figure~\ref{fig:interface} appears twice in the
 explanation (the observations used in the explicit cognitive chunks
 appear only once, but the final chunk appears twice).
\end{itemize}

\subsection{Metrics}
In a real domain, one may have a very specific metric, such as false
positive rate given no more than two minutes of thinking or the number
of publishable scientific insights enabled by the explanation. In our
experiments, we considered three basic metrics that are likely to be
relevant to most tasks: response time, accuracy, and subjective
satisfaction:
\begin{itemize}
\item \textbf{Response Time} was measured as the number of seconds
 from when the task was displayed until the subject hit the submit
 button on the interface.
\item \textbf{Accuracy} was measured if the subject correctly
 identified output consistency for verification questions, the
 presence or absence of a change in recommendation correctness 
 under the perturbation for counterfactual questions, and 
 any combination of correct categories for simulation questions.
\item \textbf{Subjective Satisfaction} was measured on a 5-point
 Likert scale. After submitting their answer for each question, but
 before being told if their answer was correct, the participant was
 asked to subjectively rate the quality of the explanation on a scale
 from 1 to 5 where 1 was very easy to use, 3 was neutral and 5 was
 very hard to use.
\end{itemize}

\subsection{Experimental Design and Interface}

The three kinds of variation and two domains resulted in six total
experiments. The experiments had parallel structures across the domains. 
For each question, we ask a simulation, a verification and 
a counterfactual version with parallel logic structures but different 
observations. Question order was block-randomized for
every participant so participants were always shown a verification,
then a simulation, then a counterfactual question, but which condition 
these came from was randomly determined. 
Domain and experiment were between subject variables, while task and explanation 
variation are within subjects variables. Each participant completed a single, 
full experiment.

\paragraph{Conditions}
The experiment levels were as follows: 
\begin{itemize}
 
\item \textbf{V1: Explanation Size.} We manipulated the length of
 the explanation (varying between 2, 5, and 10 lines) and the length
 of the output clause (varying between 2 and 5 terms). Each
 combination was tested once with each question type, for a total of 18
 questions. All inputs appeared 3 times in the decision set.

\item \textbf{V2: Cognitive Chunks.} We manipulated the number of
 cognitive chunks (repeated clauses in disjunctive normal form of the inputs) 
 introduced (varying between 1, 3 and 5), and whether they
 were embedded into the explanation (implicit) or abstracted out into new
 cognitive chunks and later referenced by name (explicit). Each combination was tested once with each question type, 
 for a total of 18 questions. 1 input was used in each experiment to evaluate 
 the answer directly, and 2 were used to evaluate the cognitive chunks, which 
 was used to evaluate the answer. All of the cognitive chunks were used to 
 determine the correct answer to ensure that the participant had to traverse 
 all concepts instead of skimming for the relevant one. For explicit cognitive chunks, the input 
 used to evaluate it appeared only once, but the chunk appeared 2 times. All 
 other inputs appeared twice. All decision sets had 4 lines in addition to 
 any explicit cognitive chunks. All output clauses had 3 elements.

\item \textbf{V3: Repeated Terms.} We manipulated the number of times the
 input conditions appeared in the explanation (varying between 2, 3, 4 and 5)
 and held the number of lines and length of clauses constant. Each
 combination was tested once within an experiment, for a total of 12
 questions. All decision sets had 6 lines. All output clauses 
 had 3 elements.

\end{itemize} 

\begin{figure}[H]
 \centering
 \includegraphics[width=.8\textwidth]{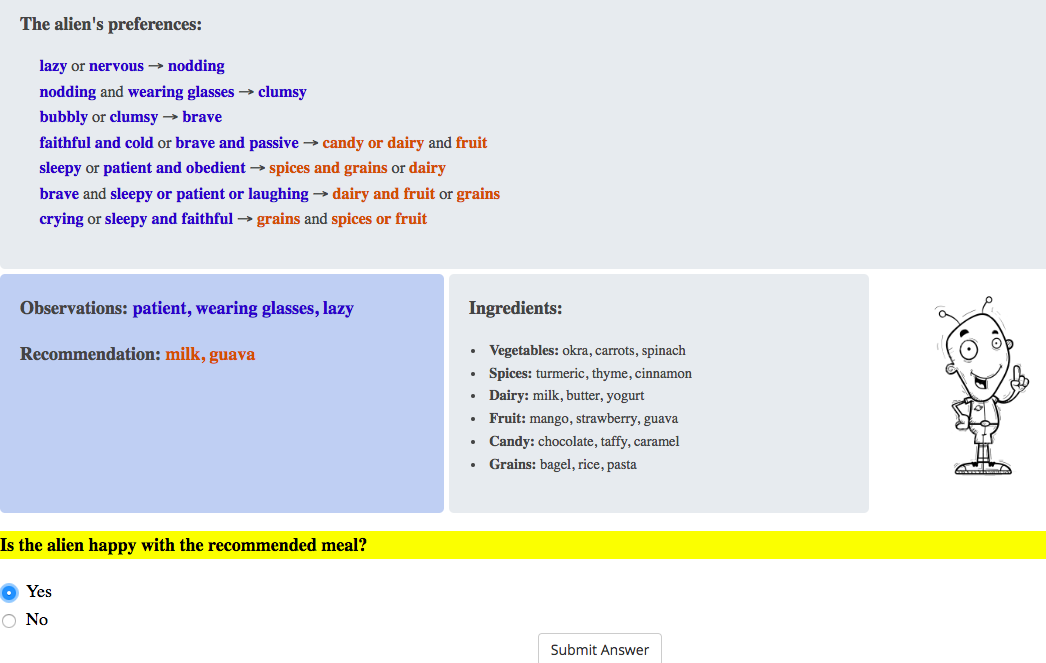}
 \caption{Screenshot of our interface for the verification task in the recipe domain. The bottom left box shows the observations we give participants about the alien, and a meal recommendation. They must then say whether the machine learning system agrees with the recommendation based on the explanation. Each task is coded in a different color (e.g., yellow) to visually distinguish them.}
 \label{fig:interface}
\end{figure}

\paragraph{Experimental Interface} 
Figure~\ref{fig:interface} shows our interface for the verification 
task in the Recipe domain. 
The \emph{observations} section refers to the
inputs into the algorithm. The \emph{recommendation} section refers
to the output of the algorithm. The \emph{preferences} section
contains the explanation---the reasoning that the supposed machine
learning system used to suggest the output (i.e., recommendation)
given the input, presented as a procedure in the form of a decision
set. Finally, the \emph{ingredients} section in the Recipe domain
(and the \emph{disease medications} section in the Clinical domain)
contained a dictionary defining \emph{concepts} relevant to the
experiment (for example, the fact that bagels, rice, and pasta are all
grains). Including this list explicitly allowed us to control for the
fact that some human subjects may be more familiar with various
concepts than others. At the bottom is where the subject completes
the task: responses were submitted using a radio button for the
verification and counterfactual questions, and using check-boxes for
simulation questions.

The choice of location for these elements was chosen based on pilot
studies (described in Appendix~\ref{sec:pilot})
---while an ordering of input, explanation, output might make
more sense for an AI expert, we found that presenting the information
in the format of Figure~\ref{fig:interface} seemed to be easier for
subjects to follow in our preliminary explorations. We also found
that presenting the decision set as a decision set seemed easier to
follow than converting it into paragraphs. Finally, we colored the
input conditions in blue and outputs in orange within the
explanation. We found that this highlighting system made it easier for
participants to parse the explanations for input conditions. We also 
highlighted the text of each question type in a different color based 
on informal feedback that it was hard to differentiate between verification 
and counterfactual questions.

\paragraph{Additional Experimental Details}
In addition to the variations and settings of interest, there were
many details that had to be fixed to create consistent experiments.
In the recipe domain, we held the list of ingredients, food
categories, and possible input conditions constant. Similarly, in the
clinical domain, we held the list of symptoms, medicine categories,
and possible input conditions constant. All questions had three
observations, and required using each one to
determine the correct answer.

Verification questions had two recommendations from two distinct
categories. Each recommendation in the verification task matched half
of the lines in the decision set. Determining the correct answer for
verification questions never required differentiating between OR and
XOR.

Counterfactual questions required a perturbation: we always perturbed
exactly one observation. The perturbed observation appeared once. The
perturbed example always evaluated to a new line of the decision set.
Counterfactual questions had a balanced number of each sequence of
true/false answers for the original and perturbed input.

Finally, to avoid participants building priors on the number of true
and false answers, the verification and counterfactual questions had
an equal number of each. This notion of balancing
did not apply to the simulation task, but in the results, high
accuracies rule out random guessing.

\paragraph{Participants}

We recruited 150 subjects for each of our six experiments 
through Amazon Mechanical Turk (900 subjects all together). 
Participants were given a tutorial 
on each task and the interface, and were told that their primary 
goal was accuracy, and their secondary goal was speed. Before completing 
the task, participants were given a set of three practice
questions, one drawn from each question type. If they answered these
correctly, they could move directly to the experiment, and otherwise
they were given an additional set of three practice questions. 

We excluded participants from the analysis who did not get all of one of the two sets 
of three practice questions correct. While this may have the effect
of artificially increasing the accuracy rates overall---we are only
including participants who could already perform the task to a
reasonable extent---this criterion helped filter the substantial
proportion of participants who were simply breezing through the
experiment to get their payment. We also excluded 6 participants who got 
sufficient practice questions correct but then took more than five 
minutes to answer a single question under the assumption that they 
got distracted while taking the experiment. Table~\ref{tab:counts} describes
the total number of participants that remained in each experiment out
of the original 150 participants.

\begin{table}[H]
 \centering
 \begin{tabular}{lrr}
  \multicolumn{3}{c}{\textbf{Number of Participants}}\\
 \multicolumn{1}{l}{Experiment} & \multicolumn{1}{c}{Recipe} & \multicolumn{1}{c}{Clinical} \\ \hline
  Explanation Size (V1) & 59 & 69 \\ 
  New Cognitive Chunks (V2) & 62 & 55 \\
  Variable Repetition (V3) & 70 & 52 \\
 \end{tabular}
 \caption{Number of participants who met our inclusion criteria for
  each experiment.}
 \label{tab:counts} 
\end{table}

Most participants were from the US or Canada 
and were less than 50 years old. A
majority had a Bachelor's degree. There were somewhat more male
participants than female. We note that US and Canadian participants
with moderate to high education dominate this survey, and results may
be different for people from different cultures and backgrounds. 
Table~\ref{tab:demographics} summarizes the demographics of all 
subjects included in the analysis across the experiments. 

\begin{table}[H]
 \centering
 \begin{tabular}{llrlrlr}
  \multicolumn{7}{c}{\textbf{Participant Demographics}}\\
  \multicolumn{1}{l}{Feature} & \multicolumn{6}{c}{Categories and Proportion} \\\hline
  \textbf{Age} & 18-34 & 61.1\% & 35-50 & 33.5\% & 51-69 & 5.4\% \\
  \textbf{Gender} & Male & 62.7\% & Female & 37.0\% & Other & 0.3\% \\
  \textbf{Education} & High School & 33.2\% & Bachelors & 53.4\% & Masters and Beyond & 9.1\% \\
  \textbf{Region} & US/Canada & 92.8\% & Asia & 4.8\% & South America & 1.6\% \\
 \end{tabular}
 \caption{Participant Demographics. There were no participants over 69 years old. 4.3\% of participants reported ``other'' for their education level. The rates of participants from Australia, Europe and Latin America were all less than 0.5\%. (All participants were included in the analyses, but we do not list specific proportions for them for brevity.)}
 \label{tab:demographics}
\end{table}

Since the exclusion criteria for the different experiments were domain specific--the question structure was the same but the words were either clinical or recipe-related, we tested that the population of participants in the recipe and clinical experiments were not significantly different. We did not test for differences between experiments in the same domain because the exclusion criteria were identical. We ran a Chi-squared test of independence (implemented in scipy.stats with a Yates continuity correction) for each demographic variable. We included only those values of the demographic variable that had at least 5 observations in both experimental populations. We found no significant differences between the proportions of any of the demographic variables.

For age, we compared proportions for: [18-34, 35-50, 51-69], and the results are: (Chi-squared test of independence, $\chi^2$=2.14, n1=177, n2=196, P=0.34, DOF=2). For gender, we compared proportions for: [Female, Male] and the results are: (Chi-squared test of independence, $\chi^2$=0.16, n1=177, n2=195, P=0.69, DOF=1). For education, we compared proportions for: [Bachelor's, High School, Masters and Beyond, Other] and the results are: (Chi-squared test of independence, $\chi^2$=0.81, n1=177, n2=196, P=0.85, DOF=3). For nationality, we compared proportions for: [Asia, US/Canada] and the results are: (Chi-squared test of independence, $\chi^2=0.23$, n1=172, n2=192, P=0.63, DOF=1).

\section{Results}

\begin{figure}[H]
\centering
\textbf{Response Time}\par\medskip
\begin{tabular}{c|c|c}
\subcaptionbox{Recipe\_V1 Time \label{2}}{\includegraphics[width = 2.10in]{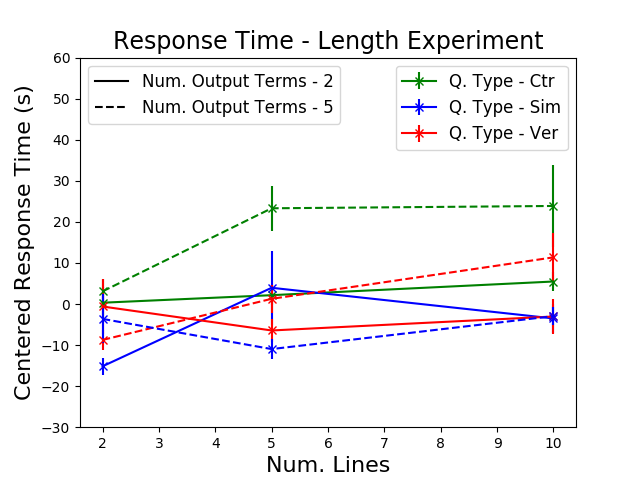}} &
\subcaptionbox{Recipe\_V2 Time \label{1}}{\includegraphics[width = 2.10in]{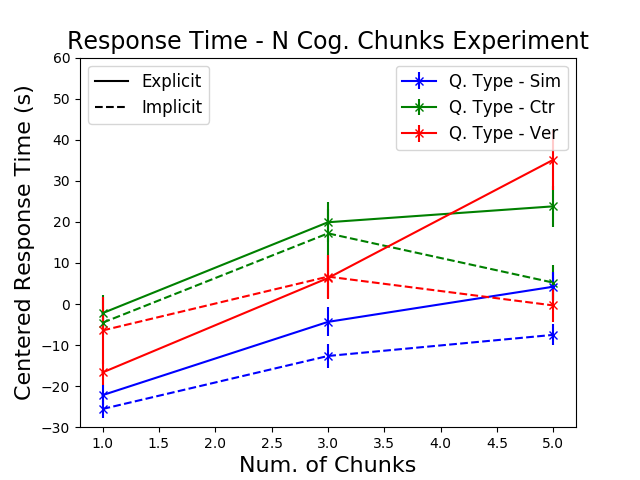}} &
\subcaptionbox{Recipe\_V3 Time\label{1}}{\includegraphics[width = 2.10in]{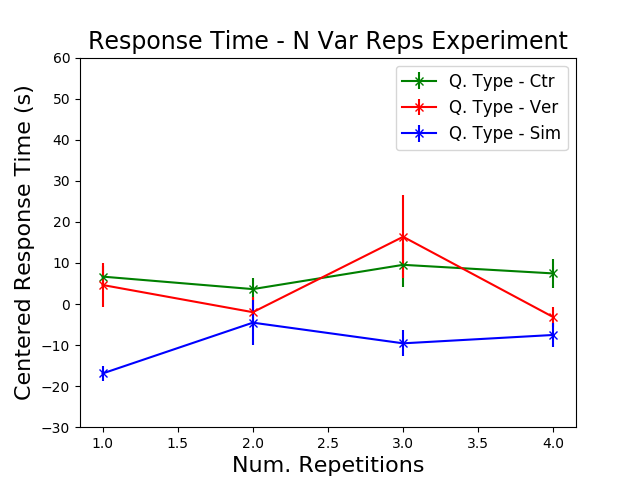}} \\
\midrule
\subcaptionbox{Clinical\_V1 Time\label{2}}{\includegraphics[width = 2.10in]{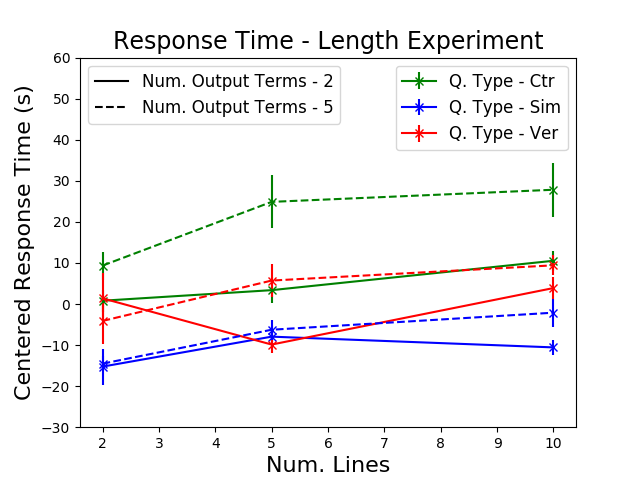}} &
\subcaptionbox{Clinical\_V2 Time\label{1}}{\includegraphics[width = 2.10in]{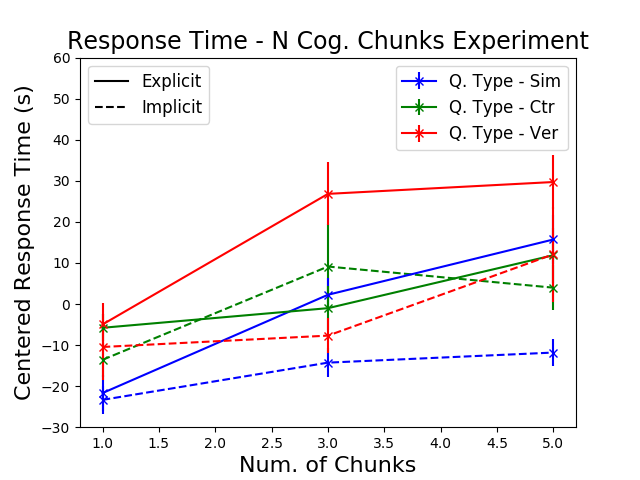}} &
\subcaptionbox{Clinical\_V3 Time\label{1}}{\includegraphics[width = 2.10in]{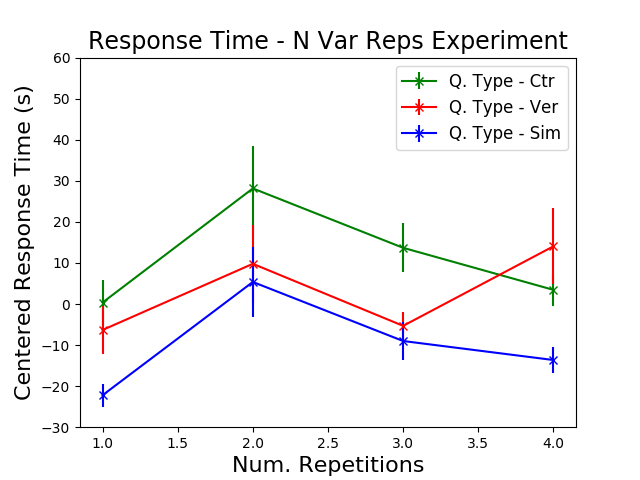}} \\
\end{tabular}
\caption{Response times across the six experiments. Responses were normalized by subtracting out the subject-specific mean to create centered values, and only response times for those subjects who got the question right are included. Vertical lines indicate standard errors.} 
\label{fig:time}
\begin{tabular}{lrrcrr}
 & \multicolumn{2}{c}{Clinical} & & \multicolumn{2}{c}{Recipe} \\
 Factor & Weight & P-Value & & Weight & P-Value \\ \hline
 Number of Lines (V1) & 1.17 & \textbf{1.41E-05} & & 1.01 & 0.00317\\
 Number of Output Terms (V1) & 2.35 & \textbf{7.37E-05} & & 1.57 & 0.0378\\
 Verification (V1) & 10.5 & \textbf{3.98E-07} & & 4.11 & 0.121\\
 Counterfactual (V1) & 21 & \textbf{1.58E-19} & & 13.7 & \textbf{1.79E-06}\\
 \\
 Number of Cognitive Chunks (V2) & 6.04 & \textbf{8.22E-11} & & 5.88 & \textbf{4.45E-17}\\
 Implicit Cognitive Chunks (V2) & -13 & \textbf{2.16E-05} & & -7.93 & \textbf{0.000489}\\
 Verification (V2) & 16.3 & \textbf{6.91E-06} & & 15.4 & \textbf{1.27E-08}\\
 Counterfactual (V2) & 8.56 & 0.0265 & & 19.9 & \textbf{6.65E-12}\\
 \\
 Number of Variable Repetitions (V3) & 1.9 & 0.247 & & 0.884 & 0.463\\ 
 Verification (V3) & 13 & 0.00348 & & 13.7 & \textbf{3.03E-05}\\ 
 Counterfactual (V3) & 20.3 & \textbf{2.41E-05} & & 16.6 & \textbf{1.91E-06}\\
\end{tabular}
 \caption{Significance tests for each factor for normalized response time. 
 A single linear regression was computed for each of V1, V2, 
 and V3. Coefficients for verification and counterfactual tasks should be 
 interpreted with respect to the simulation task. Highlighted p-values are significant at $\alpha$ = 0.05
 with a Bonferroni multiple comparisons correction across all tests
 of all experiments.}
 \centering 
\label{tab:time-pvalues}
\end{figure}

\begin{figure}[H]
\centering
\textbf{Accuracy}\par\medskip
\begin{tabular}{c|c|c}
\subcaptionbox{Recipe\_V1 Accuracy \label{2}}{\includegraphics[width = 2.10in]{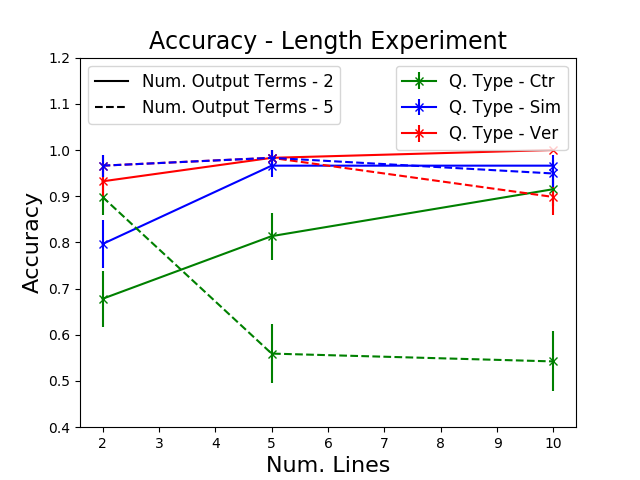}} &
\subcaptionbox{Recipe\_V2 Accuracy \label{1}}{\includegraphics[width = 2.10in]{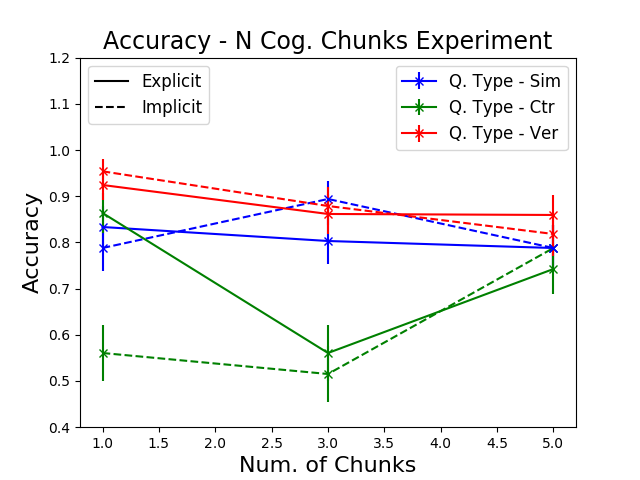}} &
\subcaptionbox{Recipe\_V3 Accuracy\label{1}}{\includegraphics[width = 2.10in]{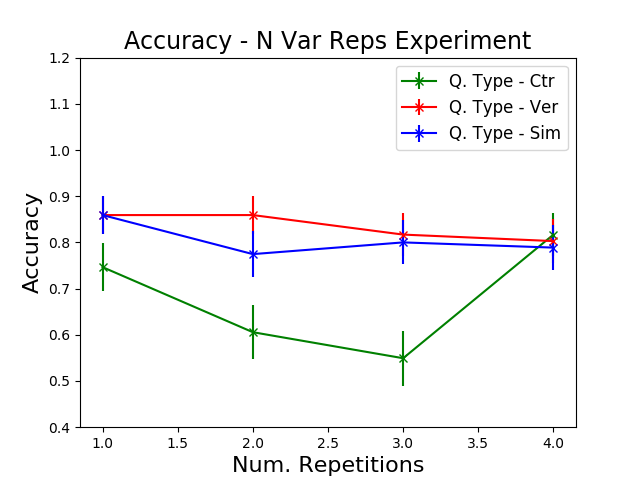}} \\
\midrule
\subcaptionbox{Clinical\_V1 Accuracy\label{2}}{\includegraphics[width = 2.10in]{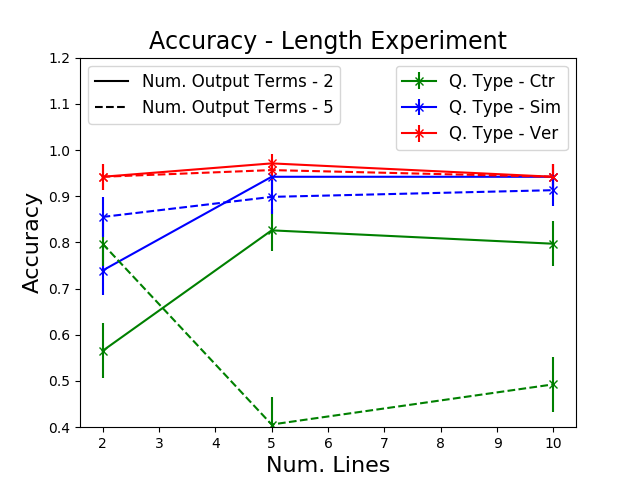}} &
\subcaptionbox{Clinical\_V2 Accuracy\label{1}}{\includegraphics[width = 2.10in]{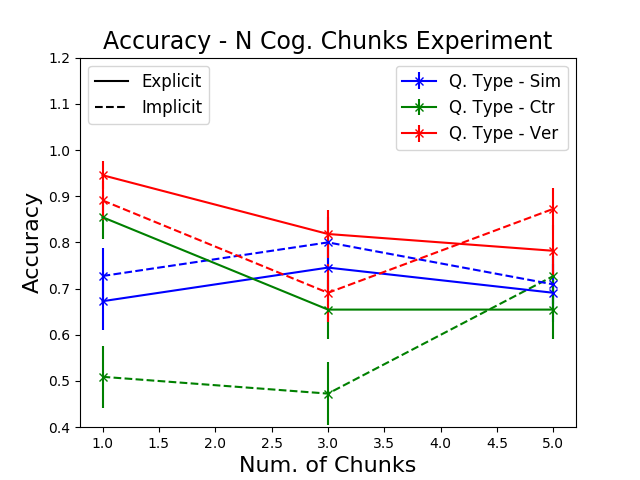}} &
\subcaptionbox{Clinical\_V3 Accuracy\label{1}}{\includegraphics[width = 2.10in]{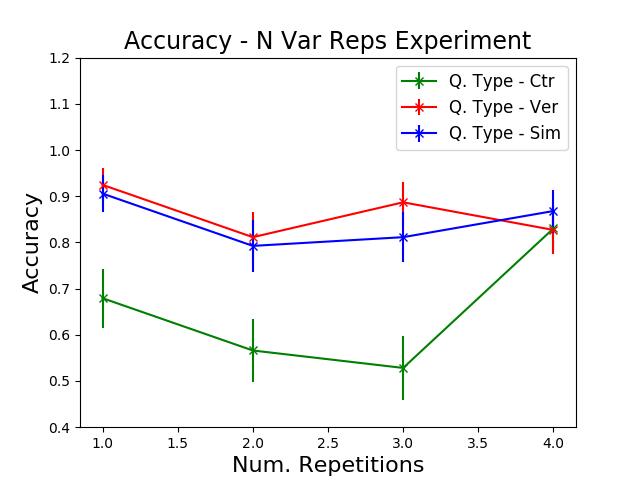}} \\
\end{tabular}
\caption{Accuracy across the six experiments. Vertical lines indicate standard errors.}
\label{fig:accuracy}
\begin{tabular}{lrrcrr}
 & \multicolumn{2}{c}{Clinical} & & \multicolumn{2}{c}{Recipe} \\
 Factor & Weight & P-Value & & Weight & P-Value \\ \hline
 Number of Lines (V1) & 0.029 & 0.236 & & 0.00598 & 0.842\\ 
 Number of Output Terms (V1) & -0.136 & 0.011 & & -0.117 & 0.0771\\
 Verification (V1) & 0.925 & \textbf{0.000652} & & 0.476 & 0.174\\
 Counterfactual (V1) & -1.41 & \textbf{1.92E-14} & & -1.7 & \textbf{1.24E-11}\\
 \\
 Number of Cognitive Chunks (V2) & -0.0362 & 0.42 & & -0.0364 & 0.416\\
 Implicit Cognitive Chunks (V2) & -0.246 & 0.093 & & -0.179 & 0.222\\
 Verification (V2) & 0.646 & 0.0008 & & 0.532 & 0.00904\\
 Counterfactual (V2) & -0.368 & 0.0294 & & -0.773 & \textbf{4.36E-06}\\
 \\
 Number of Variable Repetitions (V3) & 0.0221 & 0.804 & & -0.0473 & 0.524\\ 
 Verification (V3) & 0.146 & 0.596 & & 0.196 & 0.371\\ 
 Counterfactual (V3) & -1.07 & \textbf{7.25E-06} & & -0.67 & \textbf{0.00066}\\
\end{tabular}
\caption{Significance tests for each factor for accuracy. 
 A single logistic regression was computed for each of V1, V2, 
 and V3. Coefficients for verification and counterfactual tasks should be 
 interpreted with respect to the simulation task. 
 Highlighted p-values are significant at $\alpha$ = 0.05
 with a Bonferroni multiple comparisons correction across all tests
 of all experiments.}
\label{tab:accuracy-pvalues}
\end{figure}

\begin{figure}[H]
\centering
\textbf{Subjective Satisfaction}\par\medskip
\begin{tabular}{c|c|c}
\subcaptionbox{Recipe\_V1 Satisfaction \label{2}}{\includegraphics[width = 1.75in]{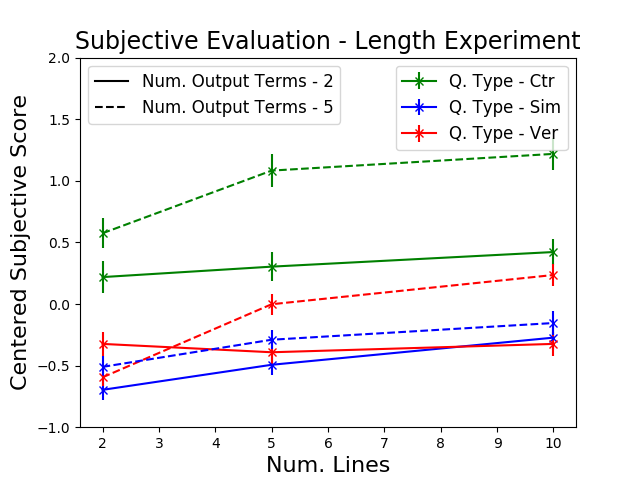}} &
\subcaptionbox{Recipe\_V2 Satisfaction \label{1}}{\includegraphics[width = 1.75in]{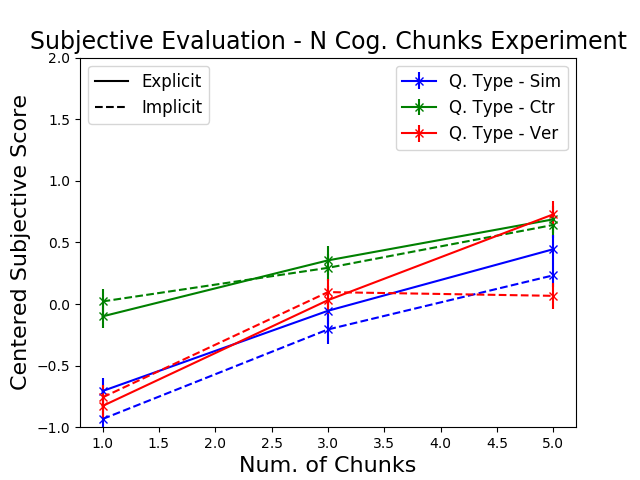}} &
\subcaptionbox{Recipe\_V3 Satisfaction \label{1}}{\includegraphics[width = 1.75in]{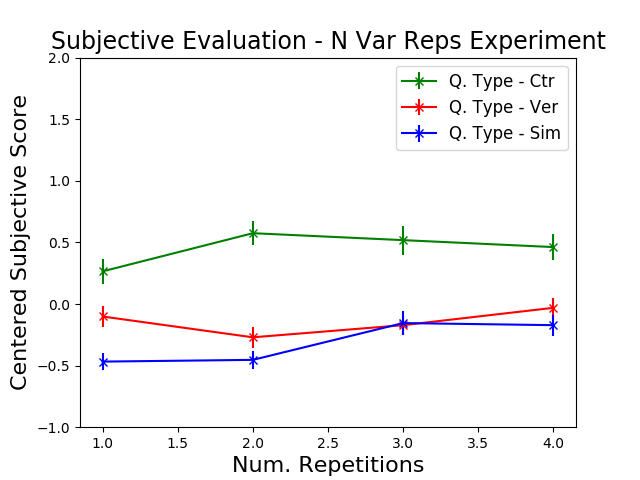}} \\
\midrule
\subcaptionbox{Clinical\_V1 Satisfaction \label{2}}{\includegraphics[width = 1.75in]{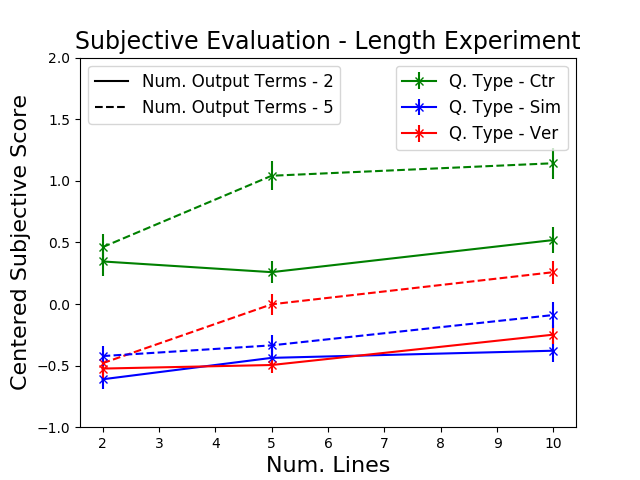}} &
\subcaptionbox{Clinical\_V2 Satisfaction \label{1}}{\includegraphics[width = 1.75in]{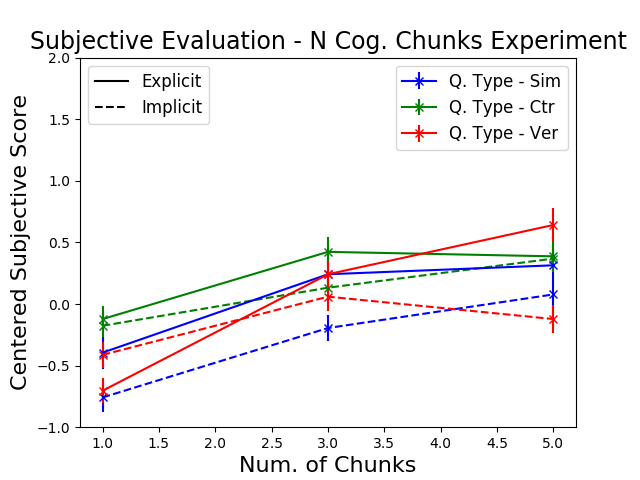}} &
\subcaptionbox{Clinical\_V3 Satisfaction \label{1}}{\includegraphics[width = 1.75in]{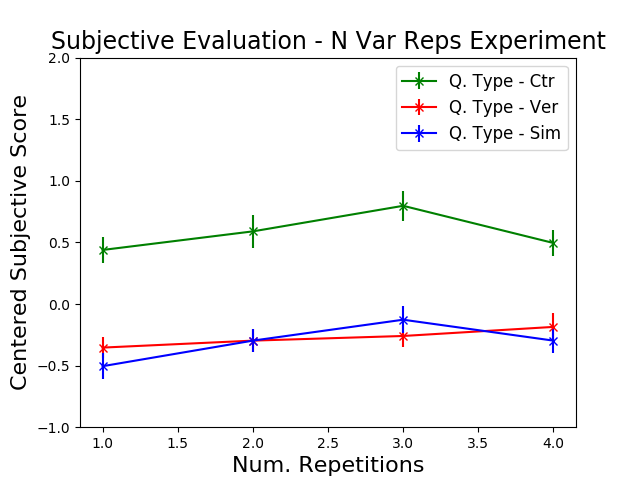}} \\
\end{tabular}
\caption{Subjective satisfaction across the six experiments. Participants were asked to rate how difficult it was to use each explanation to complete the task on a scale from 1 (very easy) to 5 (very hard). Responses were normalized by subtracting out the subject-specific mean to create centered values. Vertical lines indicate standard errors.}
\label{fig:subjective}
\begin{tabular}{lrrcrr}
 & \multicolumn{2}{c}{Clinical} & & \multicolumn{2}{c}{Recipe} \\
 Factor & Weight & P-Value & & Weight & P-Value \\ \hline
 Number of Lines (V1) & 0.0495 & \textbf{8.5E-13} & & 0.0491 & \textbf{5.57E-11}\\
 Number of Output Terms (V1) & 0.116 & \textbf{2.28E-14} & & 0.116 & \textbf{2.54E-12}\\
 Verification (V1) & 0.13 & 0.0187 & & 0.169 & 0.00475\\
 Counterfactual (V1) & 1.01 & \textbf{1.16E-65} & & 1.04 & \textbf{1.5E-59}\\
 \\
 Number of Cognitive Chunks (V2) & 0.177 & \textbf{3.96E-24} & & 0.254 & \textbf{3.76E-54}\\
 Implicit Cognitive Chunks (V2) & -0.228 & \textbf{4.1E-05} & & -0.121 & 0.0171\\
 Verification (V2) & 0.0697 & 0.305 & & 0.092 & 0.14\\
 Counterfactual (V2) & 0.288 & \textbf{2.42E-05} & & 0.52 & \textbf{1.61E-16}\\
 \\
 Number of Variable Repetitions (V3) & 0.057 & 0.0373 & & 0.0676 & 0.00411\\ 
 Verification (V3) & 0.0326 & 0.664 & & 0.169 & 0.00899\\ 
 Counterfactual (V3) & 0.887 & \textbf{1.93E-29} & & 0.767 & \textbf{2.41E-30}\\
\end{tabular}
\caption{Significance tests for each factor for normalized subjective satisfaction. 
A single linear regression was computed for each of V1, V2, 
and V3. Coefficients for verification and counterfactual tasks should be 
 interpreted with respect to the simulation task. 
 Highlighted p-values are significant at $\alpha$ = 0.05
with a Bonferroni multiple comparisons correction across all tests
of all experiments.}
\label{tab:subjective-pvalues}
\end{figure}

We report response time, accuracy and subjective satisfaction (whether participants 
thought the task was easy to complete or not) across all six experiments in 
Figures~\ref{fig:time},~\ref{fig:accuracy} and~\ref{fig:subjective}, respectively. Response time
is shown for subjects who correctly answered the questions. 
Response time and subjective satisfaction were
normalized across participants by subtracting the participant-specific
mean.

We evaluated the statistical significance of the trends in
these figures using a linear regression for the continuous outputs
(response time, subjective score) and a logistic regression for binary
output (accuracy). For each outcome, one regression was performed for
each of the experiments V1, V2, and V3. If an experiment had more
than one independent variable---e.g., number of lines and terms in
output---we performed one regression with both variables. We included 
whether the task was a verification or a counterfactual question as 
2 distinct binary variables that should be interpreted with respect to the simulation task. 
Regressions were performed with the statsmodels library
\citep{seabold2010statsmodels} and included an intercept term. In 
Table~\ref{tab:time-pvalues}, Table~\ref{tab:accuracy-pvalues} and
Table~\ref{tab:subjective-pvalues}, we report the results of these 
regressions with p-values that are significant at $\alpha = 0.05$ after a Bonferroni multiple
comparisons correction across all tests of all experiments highlighted in bold.

We next describe our main findings. We find that greater complexity
results in longer response times, but the magnitude of the effect
varies across the different kinds of explanation variation in
sometimes unexpected ways. These results are consistent across
domains, tasks, and metrics. 

\paragraph{Greater complexity results in longer response times, 
 but the magnitude of the effect varies by the type of complexity.}
Unsurprisingly, adding complexity generally increases response times.
In Figure~\ref{fig:time}, we see that increasing the number of lines,
the number of terms within a line, adding new cognitive chunks, and repeating
variables all show trends towards increasing response time.
Table~\ref{tab:time-pvalues} reports which of these response time
trends are statistically significant: the number of cognitive chunks
and whether these are implicitly embedded in the explanation or
explicitly defined had a statistically
significant effect on response time in both domains, the number of
lines and the number of output terms had a statistically significant
effect on response time only in the recipe domain, and the number of
repeated variables did not have a statistically significant effect in
either domain.

More broadly, the magnitude of the increase in response time varies
across these factors. (Note the the y-axes in Figure~\ref{fig:time}
all have the same scale for easy comparison.) Introducing new
cognitive chunks can result in overall increases in response time on
the order of 20 seconds, whereas increases in length has increases on
the order of 10 seconds. Increases in variable repetition does not
always have a consistent effect, and even when it does, the trend does
not appear to be on the order of more than a few seconds. These
relationships have implications for designers of explanation systems:
variable repetitions seem to have significantly less burden than new
concepts.

Another interesting finding was that participants took significantly
longer to answer when new cognitive chunks were made explicit rather
than implicitly embedded in a line. That is, participants were faster
when they had to process fewer, longer lines (with an implicit
concept) rather than when they had to process more, simpler lines.
One might have expected the opposite---that is, it is better to break
complex reasoning into smaller chunks---and it would be interesting to
unpack this effect in future experiments. For example, it could be
that explicitly instantiating new concepts made the explanation harder
to scan, and perhaps highlighting the locations of input terms (to
make them easier to find) would negate this effect.

\paragraph{Consistency across domains: Magnitudes of effects change, but trends stay the same.}
In all experiments, the general trends are consistent across both the
recipe and clinical domains. Sometimes an effect is weaker or unclear,
but never is an effect clearly reversed. There were 21 cases of factors 
that had a statistically significant effect on a dependent variable in at least 
1 of the 2 domains. For 19 of those, the 95\% confidence interval of both domains 
had the same sign (i.e., the entire 95\% confidence interval was positive for both domains or negative for both domains). For the other 2 (the effect of verification questions on accuracy 
and response time for experiment V1), one of the domains (clinical) was inconclusive 
(interval overlaps zero). The consistency of the signs of the effects bodes well for there being
a set of general principles for guiding explanation design, just as
there exist design principles for user interfaces and human-computer
interaction.

\paragraph{Consistency across tasks: Relative trends stay the same, different absolute values.}
The effects of different kinds of complexity on response time were
also consistent across tasks. That said, actual response times
varied significantly between tasks. In Figure~\ref{fig:time}, we see
that the response times for simulation questions are consistently low,
and the response times for counterfactual questions are consistently
high (statistically significant across all experiments except V2 in
the Recipe domain). Response times for verification questions are
generally in between, and often statistically significantly higher
than the easiest setting of simulation. For designers of explanation,
this consistency of the relative effects of different explanation
variations bodes well for general design principles. For designers of
tasks, the differences in absolute response times suggests that the
framing of the task does matter for the user.

\paragraph{Consistency across metrics: Subjective satisfaction follows response time, less clear trends in accuracy.}
So far, we have focused our discussion on trends with respect to
response time. In Table~\ref{tab:subjective-pvalues}, we see that
subjective satisfaction largely replicates the findings of response
time. We see a statistically significant preference for simulation
questions over counterfactuals. We also see a statistically
significant effect of the number of cognitive chunks, explanation
length, and number of output terms. The finding that people prefer
implicit cognitive chunks to explicit cognitive chunks appeared only
in the recipe domain. These results suggest that, in general,
peoples' subjective preferences may reflect objective measures of
interpretability like response time. However, we must also keep in
mind that especially for a Turk study, subjective satisfaction may
match response times because faster task completion corresponds to a
higher rate of pay. 

Unlike response time and subjective satisfaction, where the trends
were significant and consistent, the effect of explanation variation
on accuracy was less clear. None of the effects due to explanation
variation were statistically significant, and there are no clear
trends. We believe that this was because in our experiments, we asked
participants to be fast but accurate, effectively pushing any effects
into response time. That said, even when participants were coached to
be accurate, some tasks proved harder than others: counterfactual
tasks had significantly lower accuracies than simulation tasks. 

\section{Discussion}
Identifying how different factors affect a human's ability to utilize
explanation is an essential piece for creating interpretable machine
learning systems---we need to know what to optimize. What factors
have the largest effect, and what kind of effect? What factors have
relatively little effect? Such knowledge can help us expand to
faithfulness of the explanation to what it is describing with minimal
sacrifices in human ability to process the explanation.

\paragraph{Consistent patterns provide guidance for design of explanation systems.}
In this work, we found consistent patterns across metrics, tasks, and
domains for the effect of different kinds of explanation variation.
These patterns suggest that, for decision sets, the introduction of
new cognitive chunks or abstractions had the greatest effect on response time,
then explanation size (overall length or length of line), and finally
there was relatively little effect due to variable repetition. These
patterns are interesting because machine learning researchers have
focused on making decision set lines orthogonal
(e.g., \citep{lakkaraju2016interpretable}), which is equivalent to
minimizing variable repetitions, but perhaps, based on these results,
efforts should be more focused on explanation length and if and how
new concepts are introduced. 

We also find consistent patterns across explanation forms for the
effect of certain tasks. Simulation was the fastest, followed by
verification and then counterfactual reasoning. The counterfactual
reasoning task also had the lowest accuracies. This suggests that
participants doing the verification and counterfactual reasoning tasks
were likely first simulating through the explanation and then doing
the verification or counterfactual reasoning. (We note that while our
results focus on response time, if participants were
time-limited, we would expect effects in response time to turn into effects
in accuracy.) While these observations are less relevant to designers
of explanation systems, they may be valuable for those considering how
to design tasks.

\paragraph{There exist many more directions to unpack.} 
While we found overall consistent and sensible trends, there are
definitely elements from this study that warrent further
investigation. Particularly unexpected was that participants had
faster response times when new cognitive chunks were implicit rather than
explicit. It would be interesting to unpack whether that effect
occurred simply because it meant one could resolve the answer in one
long line, rather than two (one to introduce the concept, one to use
it), and whether the familiarity of the new concept has an effect.
More broadly, as there are large ML efforts around representation
learning, understanding how humans understand intermediate concepts is an
important direction.

Other interesting experimental directions include the metrics and the
interface. Regarding the metrics, now that we know what kinds of
explanations can be processed the fastest, it would be interesting to
see if subjective satisfaction correlates to those variables in the
absence of a task. That is, without any time pressure or time
incentive, do people still prefer the same properties purely
subjectively? Regarding the interface, we chose ours based on
several rounds of pilot studies and then fixed it. However, one can
imagine many ways to optimize the display of information, such as
highlighting relevant lines. Ultimately, the choice of the display
will have to reflect the needs of the downstream task.

More broadly, there are many interesting directions regarding what
kinds of explanation are best in what contexts. Are there universals
that make for interpretable procedures, whether they be cast as
decision sets, decision trees, or more general pseudocode; whether the
task is verification, forward simulation, or counterfactual reasoning?
Do these universals also carry over to regression settings? Or does
each scenario have its own set of requirements? When the
dimensionality of an input gets very large, do trade-offs for defining
new intermediate concepts change? A better understanding of these
questions is critical to design systems that can provide effective
explanation to human users.

Finally, future work will need to connect performance on these basic
tasks to finding errors, deciding whether to trust the model, and
other real-world tasks. These tasks are more difficult to do in controlled
settings because each user must have a similar level of grounding in
the experimental domain to determine whether an action might be, for
example, safe. Our work is just one part of a process of building our
understanding of how humans use explanation.

\section{Conclusion}
In this work, we investigated how the ability of humans to perform a
set of simple tasks---simulation of the response, verification of a
suggested response, and determining whether the correctness of a
suggested response changes under a change to the inputs---varies as a
function of explanation size, new types of cognitive chunks and
repeated terms in the explanation. We found consistent effects
across tasks, metrics, and domains, suggesting that there may exist
some common design principles for explanation systems. 

\paragraph{Acknowledgements}
The authors acknowledge PAIR at Google, a Google Faculty Research
Award, and the Harvard Berkman Klein Center for their support. IL is
supported by NIH 5T32LM012411-02. 

\bibliographystyle{plainnat}
\bibliography{main}

\appendix 

\section*{Interface} 

\begin{figure}[h]
 \centering
 \includegraphics[width=.55\textwidth]{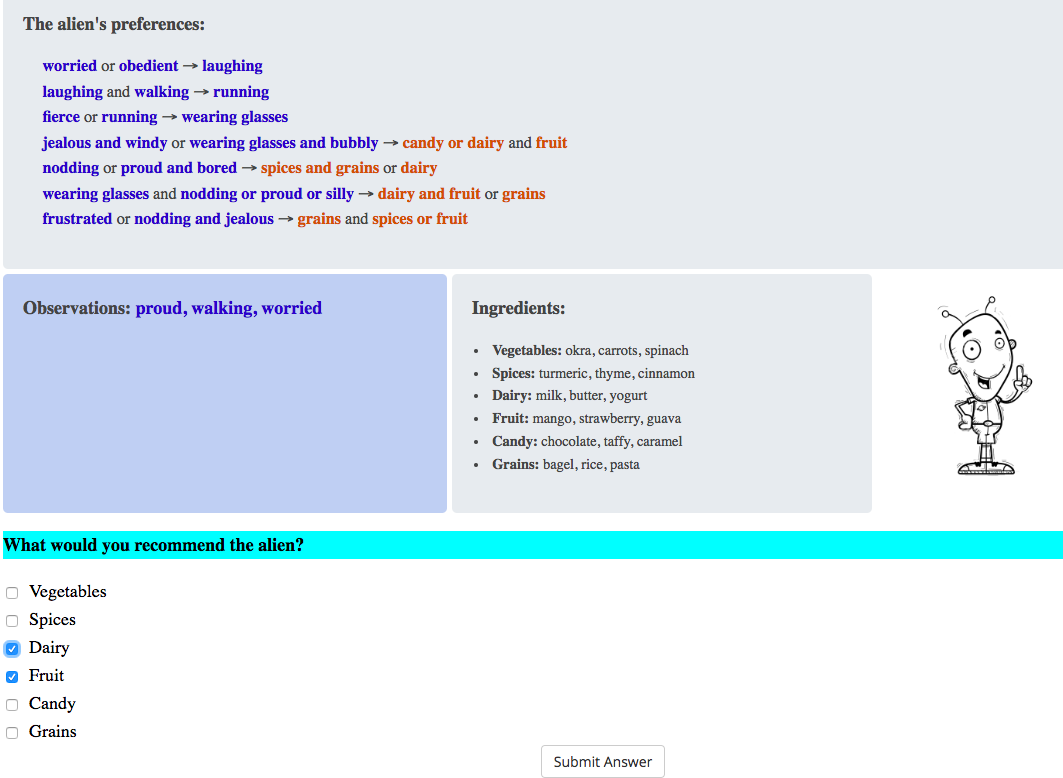}
 \caption{Screenshot of our interface for the simulation task in the Recipe domain. Participants must give a valid recommendation that will satisfy the alien given the observations and preferences.}
 \label{fig:interface_sim}
\end{figure}

\begin{figure}[h]
 \centering
 \includegraphics[width=.6\textwidth]{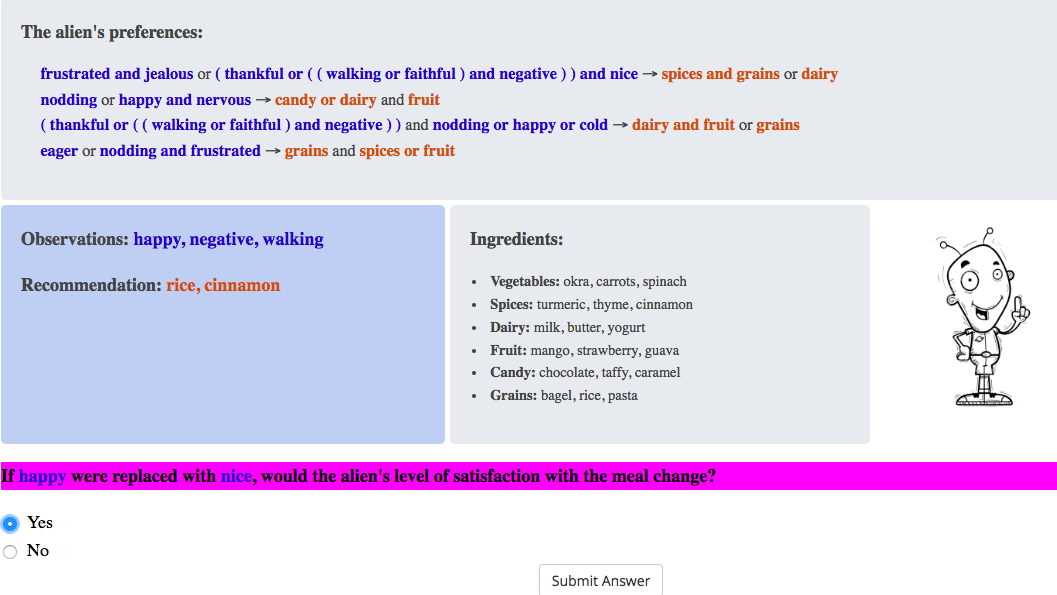}
 \caption{Screenshot of our interface for the counterfactual task in the Recipe domain. Participants must determine whether the alien's satisfaction with the recommendation changes under the change to the observations described in the magenta box given the observations and the alien's preferences.}
 \label{fig:interface_ctr}
\end{figure}

\section*{Description of Pilot Studies} 
\label{sec:pilot}

We conducted several pilot studies in the design of these experiments.
Our pilot studies showed that asking subjects to respond quickly or
within a time limit resulted in much lower accuracies; subjects would
prefer to answer as time was running out rather than risk not
answering the question. That said, there are clearly avenues of
adjusting the way in which subjects are coached to place them in fast
or careful thinking modes, to better identify which explanations are
best in each case.

The experiment interface design also played an important role. We
experimented with different placements of various blocks, the coloring
of the text, whether the explanation was presented as rules or as
narrative paragraphs, and also, within rules, whether the input was
placed before or after the conclusion (that is, `if A: B'' vs. ``B if
A''). All these affected response time and accuracy, and we picked
the configuration that had the highest accuracy and user satisfaction.

Finally, in these initial trials, we also varied more factors: number
of lines, input conjunctions, input disjunctions, output conjunctions,
output disjunctions and global variables. Upon running preliminary
regressions, we found that there was no significant difference in
effect between disjunctions and conjunctions, though the number of
lines, global variables, and general length of output
clause---regardless of whether that length came from conjunctions or
disjunctions---did have an effect on the response time. Thus, we chose
to run our experiments based on these factors.

\end{document}